\documentclass{sigchi-lex}

\RequirePackage{snapshot}

\usepackage{graphics} 
\usepackage[T1]{fontenc}
\usepackage{txfonts}
\usepackage{mathptmx}
\usepackage[pdfpagelabels=false]{hyperref}
\usepackage{color}
\usepackage[usenames,dvipsnames,table]{xcolor}
\usepackage{booktabs}
\usepackage{textcomp}
\usepackage[all]{hypcap}  
\usepackage{ccicons}  
\usepackage{float}
\usepackage{multicol}
\usepackage{cuted}

\hypersetup{%
  colorlinks,
  citecolor=black,
  filecolor=black,
  linkcolor=black,
  urlcolor=black
}

\usepackage{caption}
\DeclareCaptionType{copyrightbox} 
\usepackage{subcaption}

\usepackage[nocompress]{cite}

\usepackage{gensymb}
\usepackage{amsmath}
\usepackage{amsfonts}
\usepackage{amssymb}

\newcommand{\bx}{\boldsymbol{x}}
\newcommand{\by}{\boldsymbol{y}}
\newcommand{\citeAuth}[1]{Authors in \cite{#1}}
\newcommand{\citeauth}[1]{authors in \cite{#1}}

\newcommand{\secref}[1]{the ``\nameref{sec:#1}'' section}
\newcommand{\figref}[1]{Fig.~\ref{fig:#1}}

\newcommand{\periphurl}{\url{https://****.***.***/peripheral}}

\title{SideEye: A Generative Neural Network Based\vspace{0.1in}\\Simulator of Human Peripheral Vision\\\vspace{0.02in}}

\numberofauthors{6}

\author{
  \alignauthor{Lex Fridman\\\affaddr{MIT}}
  \alignauthor{Benedikt Jenik\\\affaddr{MIT}}
  \alignauthor{Shaiyan Keshvari\\\affaddr{MIT}}
  \alignauthor{Bryan Reimer\\\affaddr{MIT}}
  \alignauthor{Christoph Zetzsche\\\affaddr{University of Bremen}}
  \alignauthor{Ruth Rosenholtz\\\affaddr{MIT}}
}

\begin{document}

\maketitle

\begin{strip}
  \centering
  \begin{subfigure}[t]{0.46\textwidth}
    \centering
    \includegraphics[width=\textwidth]{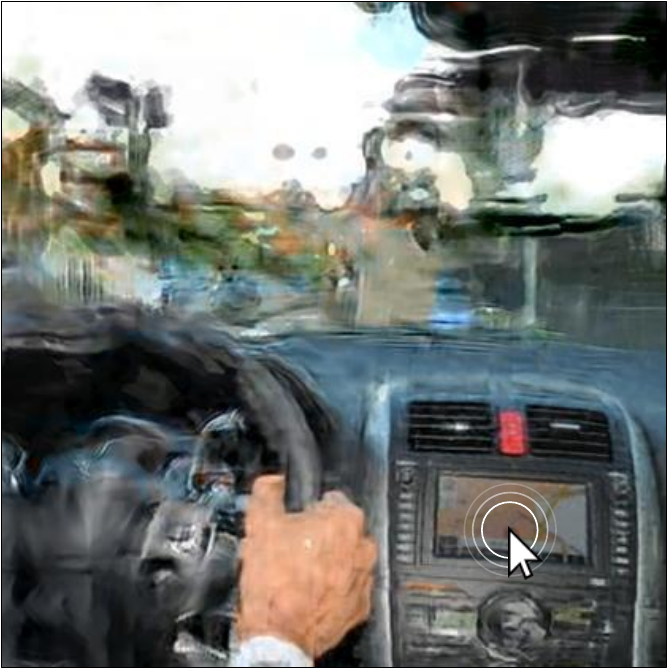}
    \caption{Looking at the GPS navigation screen in the center stack.}
    \label{fig:pedestrian-center-stack}
  \end{subfigure}\hspace{0.4in}
  \begin{subfigure}[t]{0.46\textwidth}
    \centering
    \includegraphics[width=\textwidth]{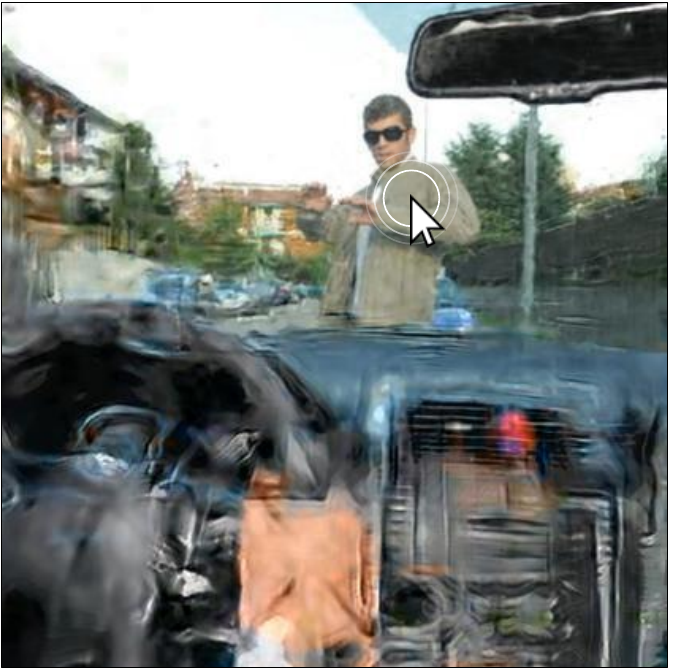}
    \caption{Looking at the pedestrian.}
  \end{subfigure}
  \captionof{figure}{Peripheral vision simulation using a generative neural network trained on a behaviorally-validated
    model of human peripheral vision. Our SideEye tool uses this model to simulate what an observer sees when foveating
    the location of the mouse pointer. Note: In the figure on the left, when looking at the center stack, the pedestrian is not
    visible in the periphery.}
  \label{fig:pedestrian}
\end{strip}

\begin{abstract}
Foveal vision makes up less than 1\% of the visual field. The other 99\% is peripheral vision. Precisely what human beings see in the periphery is both obvious and mysterious in that we see it with our own eyes but can't visualize what we see, except in controlled lab experiments. Degradation of information in the periphery is far more complex than what might be mimicked with a radial blur. Rather, behaviorally-validated models hypothesize that peripheral vision measures a large number of local texture statistics in pooling regions that overlap and grow with eccentricity. In this work, we develop a new method for peripheral vision simulation by training a generative neural network on a behaviorally-validated full-field synthesis model. By achieving a 21,000 fold reduction in running time, our approach is the first to combine realism and speed of peripheral vision simulation to a degree that provides a whole new way to approach visual design: through peripheral visualization.
\end{abstract}

\newcommand{\anex}[2]{
  \begin{subfigure}[t]{3in}
    \includegraphics[width=\textwidth]{images/examples/example3/image-#1.jpg}
    \caption{#2}
    \label{fig:example-#1}
  \end{subfigure}
}
\begin{figure*}[ht!]
  \centering
  \anex{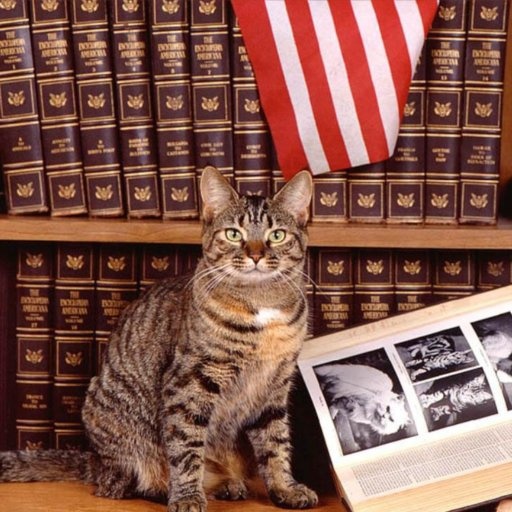}{Original $512 \times 512$ pixel image.}\hspace{0.3in}
  \anex{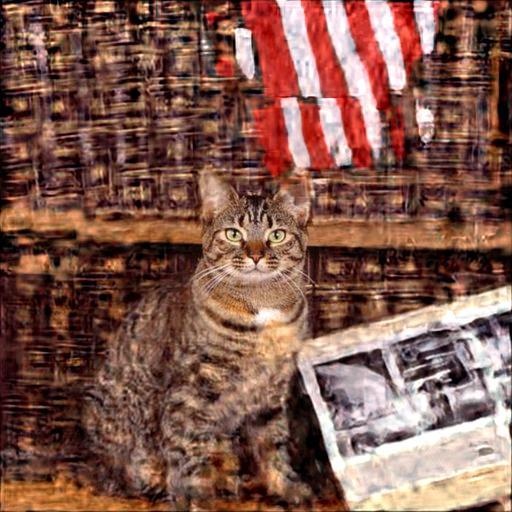}{Foveated image with fixation on $(256,256)$.}
  \caption{Example ``foveated'' image. Given an original image (a), the foveated image provides a visualization of the
    information available from a single glance (b) assuming the two subfigures are 14.2 inches away from the observer on
    a printed page (2.6 inches in height and width) and the observer's eyes are fixated at the center of the original
    image. By selecting a new model fixation, one can similarly get predictions for that fixation. Any part of the image
    that appears clearly in (b) is predicted to be easy to perceive when fixating at the center of (a). The stripes in
    the flag are clear. The cat is clearly identifiable as such, and clearly sits next to a book. The pictures within
    the book are not predicted to have a clear organization, given the modeled fixation.  The image in this example
    comes from the dataset used in the paper.}
  \label{fig:examples}
\end{figure*}

\section[Introduction]{Introduction and Related Work}\label{sec:introduction}

In the fovea (the central rod-free area of the retina, approximately 1.7\degree~in diameter), recognition is
relatively robust and effortless. However, more than 99\% of visual field lies outside the fovea, here referred to as
the periphery. Peripheral vision has considerable loss of information relative to the fovea. This begins at the retina,
which employs variable spatial resolution to get past the bottleneck of the optic nerve. However, it does not end there,
but continues with neural operations in visual cortex. Reduced peripheral acuity has only a tiny effect, compared with
peripheral vision's sensitivity to clutter, known as visual crowding (discussed in more detail in
\secref{peripheral-vision}). Unlike acuity losses, which impact only tasks relying on quite high spatial frequencies,
crowding occurs with a broad range of stimuli \cite{pelli2008uncrowded}. It is ever-present in real-world vision, in
which the visual system is faced with cluttered scenes full of objects and diverse textures. Crowding constrains what we
can perceive at a glance.

The field of human vision has recently made significant advances in understanding and modeling peripheral vision. A
successful model has been shown to predict performance at a number of peripheral and full-field vision tasks
\cite{balas2009summary,freeman2011metamers,rosenholtz2012summary,zhang2015cube,keshvari2016pooling}. Critically, from
the point of view of converting this understanding to design intuitions, researchers have visualized both reduced acuity
and visual crowding using foveated texture synthesis techniques that generate new image samples that have the same
texture statistics in a large number of overlapping ``pooling'' regions
\cite{freeman2011metamers,rosenholtz2012rethinking}. Such visualizations facilitate intuitions about peripheral vision,
e.g. for design\cite{rosenholtz2011your}, and also enable testing models of peripheral vision. However, generating each
synthesis can take a number of hours, limiting the utility of this technique.

In this work, we develop and release a Foveated Generative Network (FGN) architecture for end-to-end learning of the
foveation task and an online tool (SideEye) for real-time simulation of peripheral vision on user-submitted designs. The
primary goal of this approach is to reduce the running time of generating a human-realistic visualization of peripheral
vision from hours to milliseconds while maintaining reasonable consistency with the behaviorally validated
models. \figref{examples} shows an example visualizing the degradation of spatial information in the periphery. Being
able to perform a visualization like this in under a second has several significant applications (see list below). As
discussed in \secref{running-time}, the average running time of 700 ms could be further significantly reduced. As it
approaches 33 ms (or 30 fps), the following applications become even more feasible:

\begin{itemize}
\item \textbf{Interface Design:} Explore various graphic user interface design options with the SideEye tool on the fly
  by adjusting the fixation point and visualizing the full-field appearance of the design given the fixation point in
  near real-time (see \figref{pedestrian}). One example of this application is the A/B testing of website designs
  \cite{hanington2012universal}. An illustrative case study of this testing-based design methodology is presented in the
  \secref{ab-testing}. Communicate design intuitions and rationale to members of the design or product team.

\item \textbf{Insights into Usability and Safety:} Quickly gain intuitions about HCI issues such as whether, in an
  automotive context, a user is likely to notice obstacles (i.e., pedestrians) while engaged with a cell phone, GPS
  system, or augmented reality headset. An example of this type of exploration using the SideEye tool is shown in \figref{pedestrian}.

\item \textbf{Behavioral Evaluation of Vision Model on HCI-relevant stimuli and tasks:} The existing peripheral vision
  model has been extensively tested on a wide range of stimuli and tasks, providing confidence that the model extends to
  HCI-relevant visual tasks. Nonetheless, additional testing is always desireable. Previous model testing has utilized
  peripheral vision visualizations to generate model predictions (see, e.g. \cite{zhang2015cube}, for the standard
  methodology). For HCI tasks, this requires generating hundreds of foveated images dependent on subject fixation
  patterns. FGN can generate the needed set of foveated images on-the-fly as the subject is performing the experiment.

\item \textbf{Video Foveation:} Fast image foveation can be applied to individual frames of a video. This is an
  important step toward producing a model of peripheral vision in real-world viewing. However, there are further
  modeling challenges like accounting for peripheral encoding of motion and maintaining temporal consistency would need
  to be added to the architecture in order make video foveation a powerful tool to explore human processing of
  spatiotemporal visual information.
\end{itemize}



The main contribution of this work is to use a deep learning approach to make a model of human vision fast enough to
provide a useful design tool. This will facilitate more effective communication of visual information, better usability,
early insight into performance of a given design prior to user testing, and better communication within the design team
\cite{rosenholtz2011predictions}. To further use of this tool, we release the code and an online in-browser version at
\periphurl. We demonstrate that the resulting model successfully approximates the state-of-the-art behaviorally
validated model, and yet is 21,000 times faster.

\begin{figure*}[htp!]
  \centering
  \includegraphics[width=0.9\textwidth]{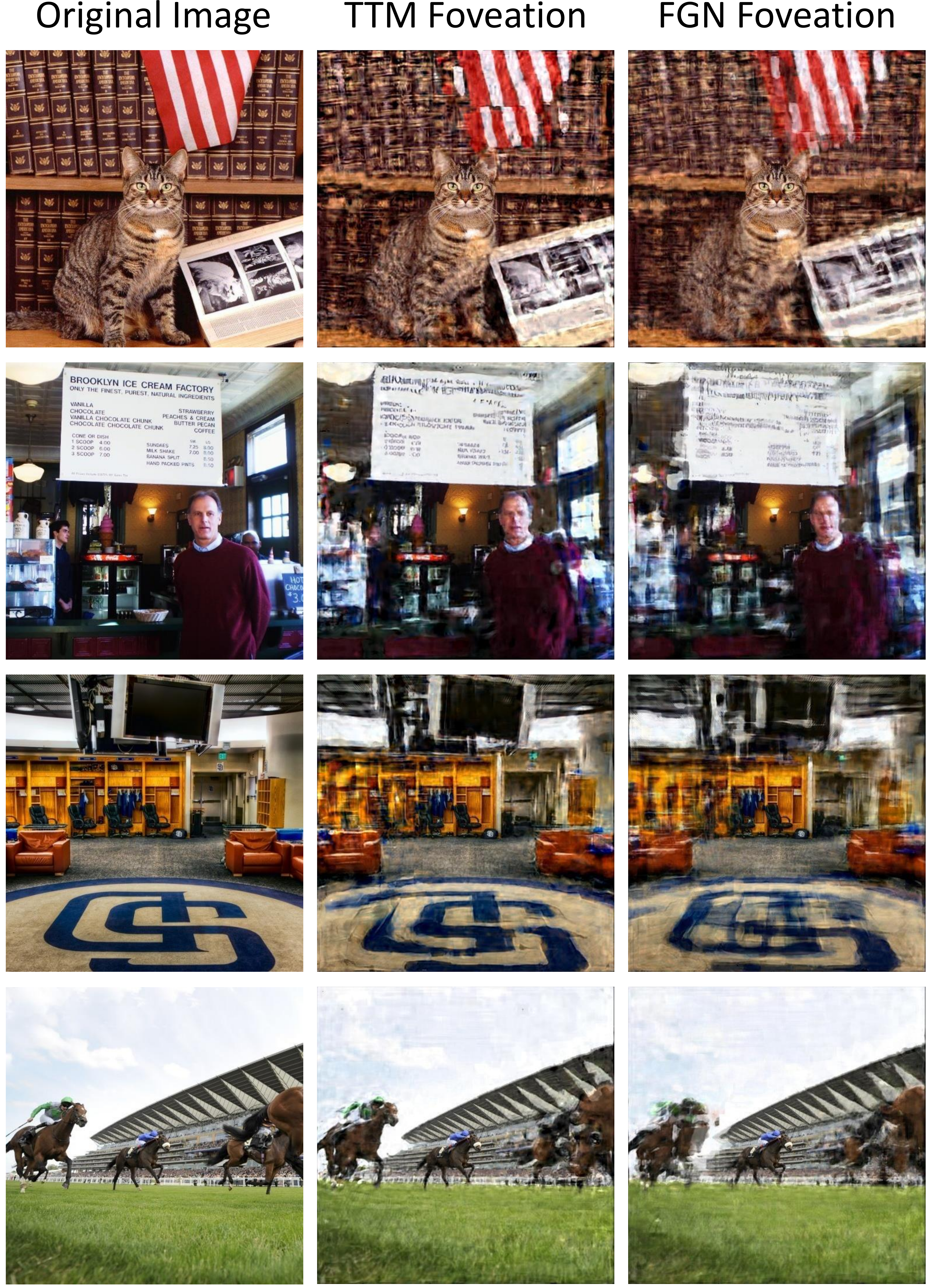}
  \caption{Four examples (rows) from the evaluation dataset showing the original image (column 1), TTM-based foveation
    of the image (column 2), and FGN-based foveation of the image (column 3). Unlike the deterministic radial blur
    function in \secref{radial-blur}, TTM is stochastic and can generate an arbitrary number of foveated images (based
    on a random number generator seed) from a single source image. Therefore, one should not expect the TTM and FGN
    foveations to match pixel by pixel, but rather their synthesis should have similar density and type of information
    degradation in the periphery.}
  \label{fig:foveations-fovea}
\end{figure*}

\section{Modeling Peripheral Vision}\label{sec:ttm}

\subsection{Crowding in Peripheral Vision}\label{sec:peripheral-vision}

\begin{figure}[h!]
  \centering
  \includegraphics[width=3in]{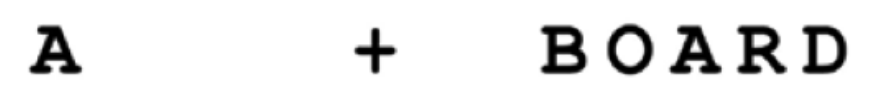}
  \caption{A classic example demonstrating the ``crowding'' effect.}
  \label{fig:aboard}
\end{figure}

It is well known that the visual system has trouble recognizing peripheral objects in the presence of nearby flanking
stimuli, a phenomenon known as crowding (for reviews see:
\cite{levi2008crowding,pelli2008uncrowded,whitney2011visual}).  \figref{aboard} shows a classic demonstration. Fixating
the central cross, one can likely easily identify the isolated `A' on the left but not the one on the right flanked by
additional letters.  An observer might see these crowded letters in the wrong order, e.g., `BORAD'. They might not see
an `A' at all, or might see strange letter-like shapes made up of a mixture of parts from several letters
\cite{lettvin1976seeing}. Move the flanking letters farther from the target `A', and at a certain critical spacing
recognition is restored. The critical spacing is approximately 0.4 to 0.5 times the eccentricity (the distance from the
center of fixation to the target) for a fairly wide range of stimuli and tasks
\cite{bouma1970interaction,pelli2009grouping,pelli2004crowding}. \citeAuth{pelli2008uncrowded} have dubbed this
\emph{Bouma's Law}.

It should be clear that such ``jumbling'' in the periphery has profound consequences for HCI design. In fact, crowding
is likely task-relevant for most real-world visual stimuli and tasks. It has a far greater impact on vision than loss of
acuity or color vision, and it is the dominant difference between foveal and peripheral vision
\cite{rosenholtz2016capabilities}. It impacts visual search, object recognition, scene perception, perceptual grouping,
shape perception, and reading \cite{rosenholtz2012summary,pelli2008uncrowded,rosenholtz2012rethinking}. Crowding
demonstrates a tradeoff in peripheral vision, in which significant information about the visual details is lost, and yet
considerable information remains to support many real-world visual tasks and to give us a rich percept of the world. The
information that survives must suffice to guide eye movements and give us a coherent view of the visual world
\cite{rolfs2011predictive}. Nonetheless, the pervasive loss of information throughout the visual field means that we
cannot hope to understand much of vision without understanding, controlling for, or otherwise accounting for the
mechanisms of visual crowding.

A fair assessment of the current state of vision research is that there exists a dominant theory of crowding.  Crowding
has been equivalently attributed to ``excessive or faulty feature integration'', ``compulsory averaging'', or ``forced
texture processing'' ``pooling'', resulting from of features over regions that grow linearly with eccentricity
\cite{lettvin1976seeing,parkes2001compulsory,levi2008crowding,pelli2008uncrowded,balas2009summary}.  Pooling has
typically been taken to mean averaging \cite{parkes2001compulsory} or otherwise computing summary statistics
\cite{lettvin1976seeing,balas2009summary} of features within the local region.

\begin{figure*}[ht!]
  \centering
  \includegraphics[width=0.7\textwidth]{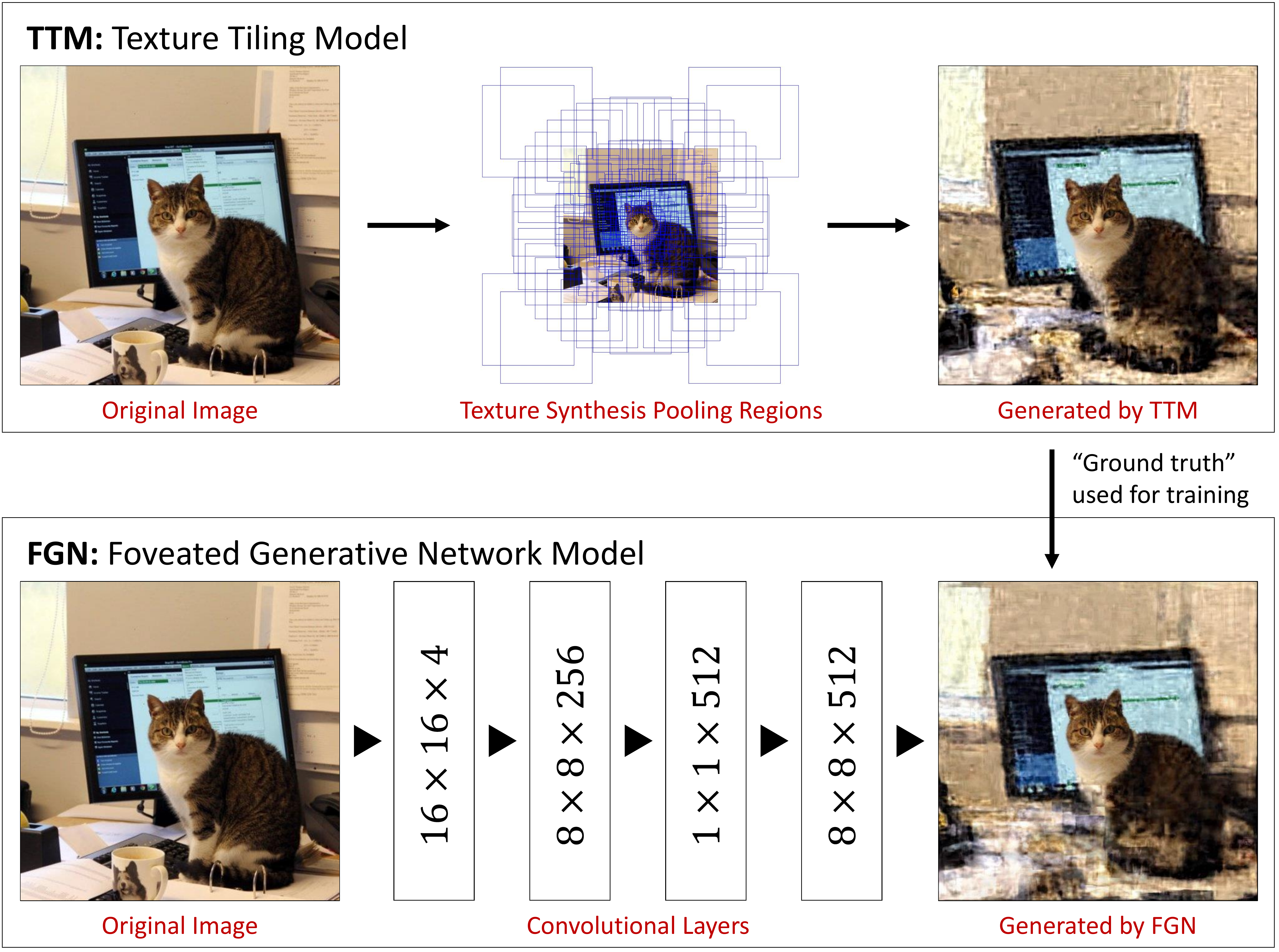}
  \caption{The architecture of the foveated generative network (FGN) used to approximate the computationally-costly
    texture tiling model (TTM). The output of the TTM is used as the ground truth for the end-to-end training of the
    FGN.}
  \label{fig:architecture}
\end{figure*}

\subsection{A Statistical Model}\label{sec:statmodel}
\citeAuth{balas2009summary} operationalized earlier theories of statistical processing in peripheral vision
\cite{lettvin1976seeing,parkes2001compulsory} in terms of measurement of a rich set of texture statistics within local
pooling regions that grow linearly with eccentricity, in accord with Bouma's Law. They used as their candidate
statistics those identified by \citeauth{portilla2000parametric}, as that set has been successful at describing texture
perception (as judged by synthesis of textures that are difficult to discriminate from the
original). \citeAuth{freeman2011metamers} generalized this model to synthesize images from local texture statistics
computed across the field of view, often referred to as the ``V2 model'', as they suggested that these computations
might occur in visual processing area V2 in the brain. \citeAuth{rosenholtz2012rethinking} have similarly developed
full-field synthesis techniques, and refer to the full model as the Texture Tiling Model (TTM).

\figref{foveations-fovea} shows four examples of foveated images generated by TTM. Note that because of the statistical
nature of the model, each input image corresponds to a large number of output images that share the same texture
statistics. Critically, these synthesized model images aid intuitions about what visual tasks will be easy to perform at
a glance, i.e. with the information available in the periphery plus the high resolution fovea. Regions of the
synthesized image that appear clear are well represented by peripheral vision, according to the model. Tasks that are
easy to perform with the synthesized images will be easy to perform at a glance. For instance, the model predicts that
it is obvious at a glance that the first image in the figure is of a cat, sitting amongst books and below a
flag. However, the layout and content of the open book may be difficult to discern.

The V2 model and TTM
\cite{freeman2011metamers,rosenholtz2012rethinking} share many features, including the local texture statistics, and
overlapping pooling regions that overlap and grow linearly with eccentricity. This paper utilizes the TTM synthesis
procedure, so we adopt that terminology.

Mounting evidence supports TTM as a good candidate model for the peripheral encoding underlying crowding; it predicts
human performance at peripheral recognition tasks
\cite{balas2009summary,freeman2011metamers,rosenholtz2012rethinking,keshvari2016pooling}, visual search
\cite{rosenholtz2012summary,zhang2015cube}, and scene perception tasks \cite{rosenholtz2012rethinking}, and equating
those local statistics creates visual metamers \cite{freeman2011metamers}.

Both the V2 model and TTM are slow to converge, as they must optimize to satisfy a large number of constraints arising
from the measured local texture statistics. \citeAuth{koenderink2012space}, on the other hand, have taken a different
approach to a related problem. They apply simple image distortions, such as spatial warping, to an image, and have shown
that it is surprisingly difficult to tell that anything is wrong away from the fovea. Applying simple image distortions
is fast to compute; however, it is not well known what distortions best capture the information available in
peripheral vision; this distortion work is not yet as well grounded in terms of being able to predict task performance
as TTM and the V2 model. Here the aim is to use deep networks to produce distortions like those introduced by TTM in a
more computationally efficient way.

\section{A Generative Model for Foveated Rendering}

\subsection[Fully Convolutional Networks]{Fully Convolutional Networks as End-to-End Generators}

Researchers have long desired to speed up successful computer vision and image processing models. In many cases these successful models have taken an image as input, and mapped that image to an output image through a slow optimization process. For example, mapping from a noisy image to a denoised image. Recent advances in neural networks have provided a solution to speeding up some of these models, by learning the nonlinear mapping between the input and output images for a given model. Once learned, one can map from input to output using a relatively fast feed-forward network.

Fully convolution neural networks and other generative neural network models have been successfully used in computer
vision literature for image segmentation \cite{long2015fully,kang2014fully}, deblurring
\cite{sun2015learning,schuler2014learning}, denoising \cite{burger2012image}, inpainting \cite{xie2012image}, and
super-resolution \cite{dong2014learning}, artifact removal \cite{eigen2013restoring}, and general deconvolution
\cite{xu2014deep}. While many of these examples learn a mapping that removes image degradation, our work aims rather to add degradation in a way that is representative of losses in human peripheral vision. As shown in
\secref{ttm-eval}, this is a highly nonlinear function in that the mapping changes significantly with variation in both global spatial context and in local
texture statistics. 

Neural networks have begun to be applied to the problem of texture synthesis, i.e. synthesizing from an example texture
a new patch of perceptually similar texture. \citeAuth{gatys2015texture} use the 16 convolutional and 5 pooling layers
of the VGG-19 network for texture synthesis. In the context of peripheral vision, their work could be viewed as a method
for synthesizing individual pooling regions as described in \secref{ttm}. However, in addition to providing only a local
texture synthesis, these new texture synthesis techniques have not been behaviorally validated as models of peripheral
vision. Adding an behaviorally-validated, human-realistic ``attentional'' (foveation) mechanism to a generative network
is a novel contributions of our work.

\newcommand{\blex}[2]{
  \begin{subfigure}[t]{0.32\textwidth}
    \includegraphics[width=\textwidth]{images/foveations/cat-library-#1.jpg}
    \caption{#2}
    \label{fig:blur-#1}
  \end{subfigure}
}
\begin{figure*}[ht!]
  \centering
  \blex{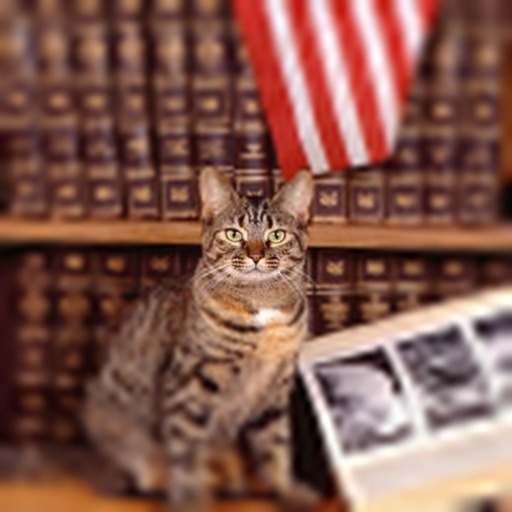}{Radial blur (ground truth).}\hspace{0.01in}
  \blex{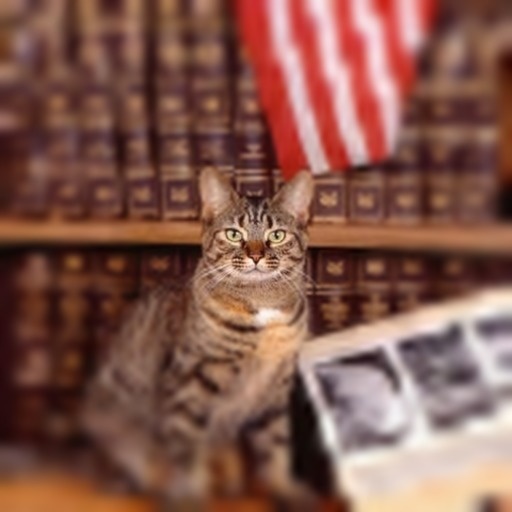}{Radial blur (FGN generated).}\hspace{0.01in}
  \blex{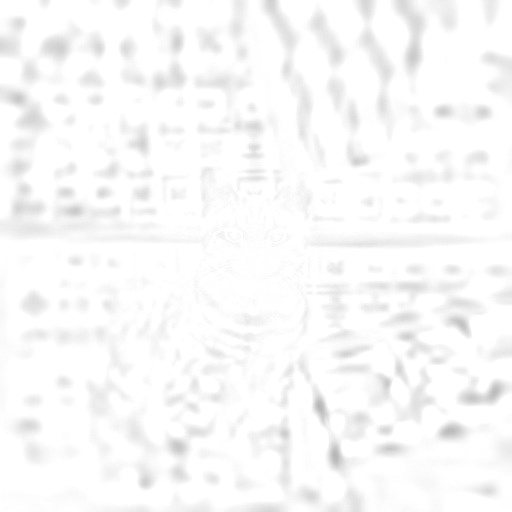}{Pixel-based difference.}
  \caption{Example output of the radial blur computed directly (a) and learned through the FGN architecture (b). The
    pixel-by-pixel difference between the two images (c) shows in black the pixels of the FGN-generated image that
    differs from the ground truth.}
  \label{fig:blur}
\end{figure*}

Given an original undistorted color image $\bx^{\{1,2,3\}}_{i,j}$ of dimension $h \times w \times 3$ and a spatial weight mask
$\bx^{\{4\}}_{i,j}$ of dimension $h \times w \times 1$, the task is to produce a foveated image, $\by$, of
dimension $h \times w \times 3$. The fourth channel of $\bx$ captures the global spatial component of the
function to be learned as it relates to the fixation point. The mask takes the form:

\begin{align*}
  d_{i,j} &= \sqrt {\left( i - f_y \right)^2 + \left( j - f_x \right)^2 }\\
  \bx_{i,j,4} & = 
                   \begin{cases}
                     d_{i,j},& \text{if } d_{i,j}> d_{\text{fovea}}\\
                     0,              & \text{otherwise}
                   \end{cases}
\end{align*}

\noindent where $d_{i,j}$ is the distance of each input pixel to the fixation point $(f_x, f_y)$, and $d_{\text{fovea}}$
is the radius (in pixels) of the foveal region. For the results in \secref{evaluation}, the fovea radius is 64 pixels.

The proposed foveated generative network (FGN) architecture is based on several components of CNN-based
deconvolution approaches \cite{burger2012image,eigen2013restoring,xu2014deep} and fully convolutional segmentation
approaches \cite{kang2014fully,long2015fully}. A fully convolutional network (FCN) can operate on large image sizes and
produce output of the same spatial dimensions. We extend the FCN architecture with the foveation weight mask (see above)
and propagate it forward through the biases of each hidden layer in order for the spatial relation with the
fixation point to be accounted for in computing the convolution and element-wise sigmoid for each layer:

\begin{equation*}
  f_k(x) = \tanh\left(w_k \cdot f_{k-1}(x) + b_k \right)
\end{equation*}

\noindent where $w_k$ and $b_k$ are the convolutions and biases at layer $k$, respectively.

In our implementation of FGN, there are 4 convolutional layers $w_{1,2,3,4}$ with $w_1$ having 256 kernels of size
$16 \times 16 \times 4$, $w_2$ having 512 kernels of size $8 \times 8 \times 256$, $w_3$ having 512 kernels of size
$1 \times 1 \times 512$, and $w_4$ having 3 kernels of size $8 \times 8 \times 512$. The loss function is defined on the
whole image pair $\left(\bx, T_i(\bx)\right)$ where $T_i(\bx)$ is the output of the TTM model on image
$\bx$ given a random seed of $i$. For purpose of FGN, this forms a unique mapping between images, but it should be
noted that TTM can generate a very large number of images $T_i(\bx), \forall i\in\mathbb{N}$ for a single input image
$\bx$, since the number of images that satisfy the statistical constraints imposed by the optimization in
TTM are upper-bounded by an exponential function in the number image pixels.

\figref{architecture} shows the fully convolutional architecture of FGN and the TTM method used to generate the
foveated image pairs. The key aspect of the former is that the foveation is completed with a single pass through the
network.

\section{Training and Evaluation of the Generative Model}\label{sec:evaluation}

We evaluate two models of foveation. The first is a naive radial blur model known to be a poor visualization of
peripheral perception as discussed in \secref{introduction}. However, it is a deterministic model for which there is a
one-to-one mapping between source image and the ground truth. Therefore, it is a good test of whether FGN can learn a
function that is spatially dependent in a global sense on the fixation point, since the pixel-wise image difference is a
more reasonable metric of comparison for a deterministic model. In addition, the greater visual interpretability of the
naive radial blur model allows us to gain intuition about the representation learning power of the FGN model. It's
important to emphasize, that the radial blur is a crude model of acuity loss that cannot serve as a reasonable model of
acuity loss in the periphery. In contrast to this, the second model of foveation we consider is the TTM model that has
been shown in behavioral experiment to capture some of the more complex characteristics of perception in the periphery
(i.e., crowding).

The original undistorted images in this paper are natural scene images selected from the Places dataset
\cite{zhou2014learning}. 1,000 images were selected for training the FGN on both the radial blur model (see
\secref{radial-blur}) and the TTM model (see \secref{ttm-eval}). Another 1,000 images were used in evaluating FGN
trained on both models. All images were cropped and resized down to $512 \times 512$ pixels. For both training and
quantitative evaluation, in this paper we assume a fixation at the center of the input image, i.e. $(w/2, h/2)$. For our
application case studies (\secref{apps}), we demonstrate the ability to move the ``fovea'' of the trained model.

\subsection[Naive Foveation]{Naive Foveation: Training on Radial Blur Model Output}\label{sec:radial-blur}

In order to evaluate the ability of FGN to learn a ``foveating'' function, we use a Gaussian blur with the standard
deviation proportional to the distance away from the fixation. The maximum standard deviation is set to 4 and decreases
linearly with distance as both approach zero. Note that this blur is made greater than that needed to mimic human
peripheral loss of acuity for the purpose of visualizing the effectiveness of our training
procedure. \figref{blur-2-blur} shows the result of applying the radial blur on one of the images from test set. This
blurring function was applied to all 1,000 images in the training set and used as $\by$ in $(\bx,\by)$ image pairs for
training an FGN network to estimate the radial blur function. \figref{blur-3-ffcn} shows the result of running the
image in \figref{example-original} through the trained network, and \figref{blur-6-diff-diff} shows the difference
between this generated image and the ground truth.

Since radial blur is a deterministic function, we can estimate the pixel error of the images generated by FGN. The
trained model was run on each of the 1,000 images in the test set and achieved an average pixel difference of 2.3. Note
that the difference shown in \figref{blur-6-diff-diff} is inverted intensity-wise for visualization clarity. This result
is a quantifiable verification that FGN can learn a simple radial blurring function, and thus presumably can capture
the loss of acuity in the periphery.

\subsection[Human-Realistic Foveation]{Human-Realistic Foveation: Training on TTM Output}\label{sec:ttm-eval}

The open question asked by this paper is whether a neural network can learn to degrade peripheral information in an
image in a way that is structurally similar to behaviorally validated models like TTM. The results shown for 4 images in
Fig. 5 and for 1,000 foveated test images made available online at \periphurl{} indicate that FGN is able to capture
many of the peripheral effects such as crowding and acuity loss. However, evaluating FGN's ability to capture the degree
of this degradation not as straightforward as evaluating a radial blur model. The TTM model produces multiple output
images for each input image, which can look radically different while still maintaining consistent texture
statistics. One does not expect the FGN output to look exactly like any given TTM output.
 Furthermore, peripheral vision loses substantial local phase (location) information, an effect well captured by
TTM. These two factors make it impossible to evaluate FGN through pixel-based comparison with the output of TTM.  We
cannot simply look at the difference between the TTM image and the FGN output, as we did when evaluating radial blur.
Here we show that FGN and TTM produce qualitatively similar distortions, and evaluate the degree to which the TTM and
FGN outputs match on the statistics explicitly measured by TTM.

In \figref{foveations-fovea}, the first column has the original images, the second column has the TTM foveated images,
and the third column has the FGN Foveation. Visual inspection of these images reveals several key observations. First,
the fovea region with the 64 pixel radius is reproduced near-perfectly (the average pixel intensity difference is below
1.9). Second, the results capture a number of known effects of crowding, including the ``jumbling'' loss of position
information, coupled with preservation of many basic features such as orientation and contrast, and dependence of
encoding quality on local image contents, including fairly good preservation of homogeneous textured regions
\cite{lettvin1976seeing,levi2008crowding,whitney2011visual}. For example, the readability of text in the periphery of
the second images is degraded significantly by its nonuniform positional displacement. Third, visual acuity decreases
with distance from the fixation point for all 4 images.

\subsection{Statistical Validation of FGN}\label{sec:validation}


\begin{figure*}[ht!]
  \centering
  \includegraphics[width=0.85\textwidth]{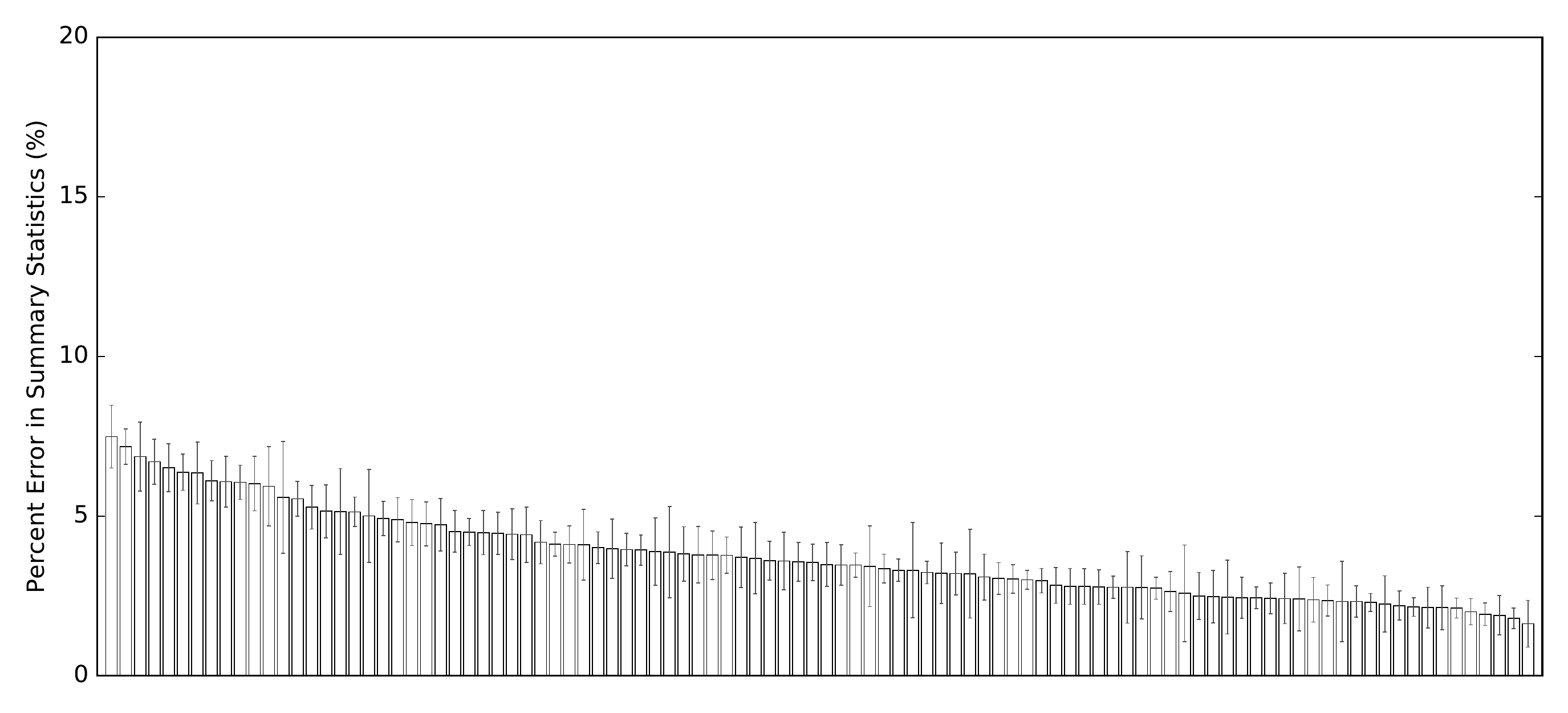}
  \caption{Percent difference (error) between FGN output and TTM output. The error is computed for each of the
    1,000 images in the test dataset and sorted from highest (left) mean error to lowest (right). Only the highest 100
    errors are shown in this figure for clarity. The mean and standard deviation of the error are computed for each
    image by aggregating over each of the values in the summary statistics vector in each of the pooling regions. All
    mean errors are below 8\%.}
  \label{fig:validation-vs-ttm}
\end{figure*}

The FGN model was trained and evaluated based on its ability to mimic, in a meaningful statistic way, the foveated
images produced by the TTM model. Therefore, statistical validation of FGN was performed by comparing its output to TTM
output over the same exact pooling regions that were used for the original TTM generation process. In other words, this
comparison evaluates the degree to which FGN is able to mimic the texture feature vector on a region by region basis and
thereby mimic the information degradation modeled by TTM. 

\figref{validation-vs-ttm}) shows the per-image difference in the feature vector representing the texture statistics in
each pooling region over that image. Each bar represents a unique image. The error in each case is computed for each of
the 1,000 images in the test dataset and sorted from highest (left) mean error to lowest (right). Only the highest 100
errors are shown in the figure for clarity. The mean and standard deviation of the error are computed for each image by
aggregating over each of the values in the summary statistics vector in each of the pooling regions. All mean errors are
below 8\% for the comparison with the TTM output.

\subsection{Running Time Performance}\label{sec:running-time}

TTM hyper-parameters were chosen such that texture synthesis convergence was achieved. For these parameters, the average
running time per image was 4.2 hours. The model is implemented in Matlab and given the structure of underlying iterative
optimization is not easily parallelizable.

The FGN architecture was implemented in TensorFlow \cite{abaditensorflow} and evaluated using NVIDIA GTX 980Ti GPU and a
2.6GHz Intel Xeon E5-2670 processor. The average running time per image was 0.7 seconds. That is an over 21,000-fold
 reduction in running time for foveating an image. There are several aspect of this performance evaluation that
indicate the possibility of significant further reductions in running time: (1) no code optimization or architecture
pruning was performed, (2) the running time includes I/O read and write operations on a SATA SSD drive, and (3) the GPU and
CPU are 2-4 years behind the top-of-the-line affordable consumer hardware.

\section{Application Case Studies}\label{sec:apps}

Peripheral vision simulation as an approach to design is enabled by this work through the release of two things: (1) an
algorithm implementation and (2) an online tool. As described in \secref{running-time}, the algorithm
implementation uses TensorFlow and Python, and provides a pre-trained generative network that takes an image as input
and produces a foveated image as output. It also takes several parameters that specify the size and position of the
fovea. These parameters control the preprocessing of the image before it is passed through the network.

The online tool (named SideEye) provides an in-browser JavaScript front-end that allows a user to load in an image that
is then passed to a Linux server backend where the FGN network run inference on the image with the fovea position in
each of 144 different locations (12x12 grid on an image). The result is 144 images, each foveated at grided
locations. When visualized together with the SideEye tool, these images automatically foveate to the position of a
hovering mouse (or, for a smartphone/tablet, the last position a finger touched the display). Two sample mouse position
for a resulting foveated set are shown in \figref{pedestrian}.

Both the SideEye tool and the pre-trained FGN network can be used to perform fast peripheral vision simulation in helping
understand how a user visually experiences the design under question when first glancing at it. In the follow 
subsections, we provide two case studies where SideEye is used to provide insight into how the design in question may
appear in the periphery and what that may mean for the overall experience that the design is intended to provide.

\subsection{Application Case Study: A/B Testing of Design Layouts}\label{sec:ab-testing}

\newcommand{\desn}[3]{
  \begin{subfigure}[t]{3.3in}
    \includegraphics[width=\textwidth]{images/design/openmile#1_#2.jpg}
    \caption{#3}
    \label{fig:design-#1-#2}
  \end{subfigure}
}
\begin{figure*}[ht!]
  \centering
  \desn{1}{original}{Old design snapshot (original image).}\hspace{0.3in}
  \desn{1}{ffcn}{Old design snapshot (image foveated on ``S'' in ``Serving'').}\\\vspace{0.1in}
  \desn{2}{original}{New design snapshot (original image).}\hspace{0.3in}
  \desn{2}{ffcn}{New design snapshot (image foveated on ``S'' in ``Serving'').}
  \caption{Given original images (a, c), the foveated images provides a visualization of
    human-realistic perception of the information available from a single glance (b, d). This particular visualization
    assumes that the two subfigures are 14.2 inches away from the observer on a printed page (2.6 inches in height and
    width) and the observer's eyes are fixated on the ``S'' in ``Serving'' (middle left). }
  \label{fig:ab-testing}
\end{figure*}

A case study of a shipping website describes a significant increase in customers requesting quotes \cite{chawla2013case}
based on a redesign shown in \figref{ab-testing}. During the design process, FGN could have been used to reveal the
information available in a glance at the website. We modeled the perception of the page when the user points his eyes at
the first word (``Serving'') of the most relevant content. Based on the output of FGN on the first design
(\figref{ab-testing}), the model predicts that a first-time visitor may be able to recognize a truck, localize the logo,
and tell that the background image contains a natural scene of some sort. Beyond that, the user might notice a black
region, with blue below it, and an orange circle. The black region, at least, appears to contain some text. Furthermore,
the model predicts that the user should be able to distinguish a light rectangle at the top of the page, and a darker
rectangle across the middle of the page, both of which the user may guess are menu bars. This is all useful information
for a user to obtain at a glance.

In the second version of the page, the model predicts that a user will easily determine that there is text against the
blue background. The location of the corporate logo is now very clear, as are both of the menu bars. The blue region in
the top half of the page contains a red rectangle, possibly containing some text. Higher level knowledge about web pages
likely suggests that this is probably a button.

Given these FGN outputs, a designer could conclude that the button in both designs is ``salient'' in the sense of being
an attention-getting color. However, the button is more obviously a button in the second design, even when fixating
several degrees away. Given these predictions, it is perhaps no surprise that users more frequently clicked on the
button to request a quote when interacting with the second design. Visualization of human perception, coupled with this
kind of reasoning process may help a designer to arriving at the more effective design prior to testing
with users, as described in \cite{chawla2013case}.

\subsection{Application Case Study: Logo Design Analysis}

\begin{figure*}[ht!]
  \centering
  \includegraphics[width=0.92\textwidth]{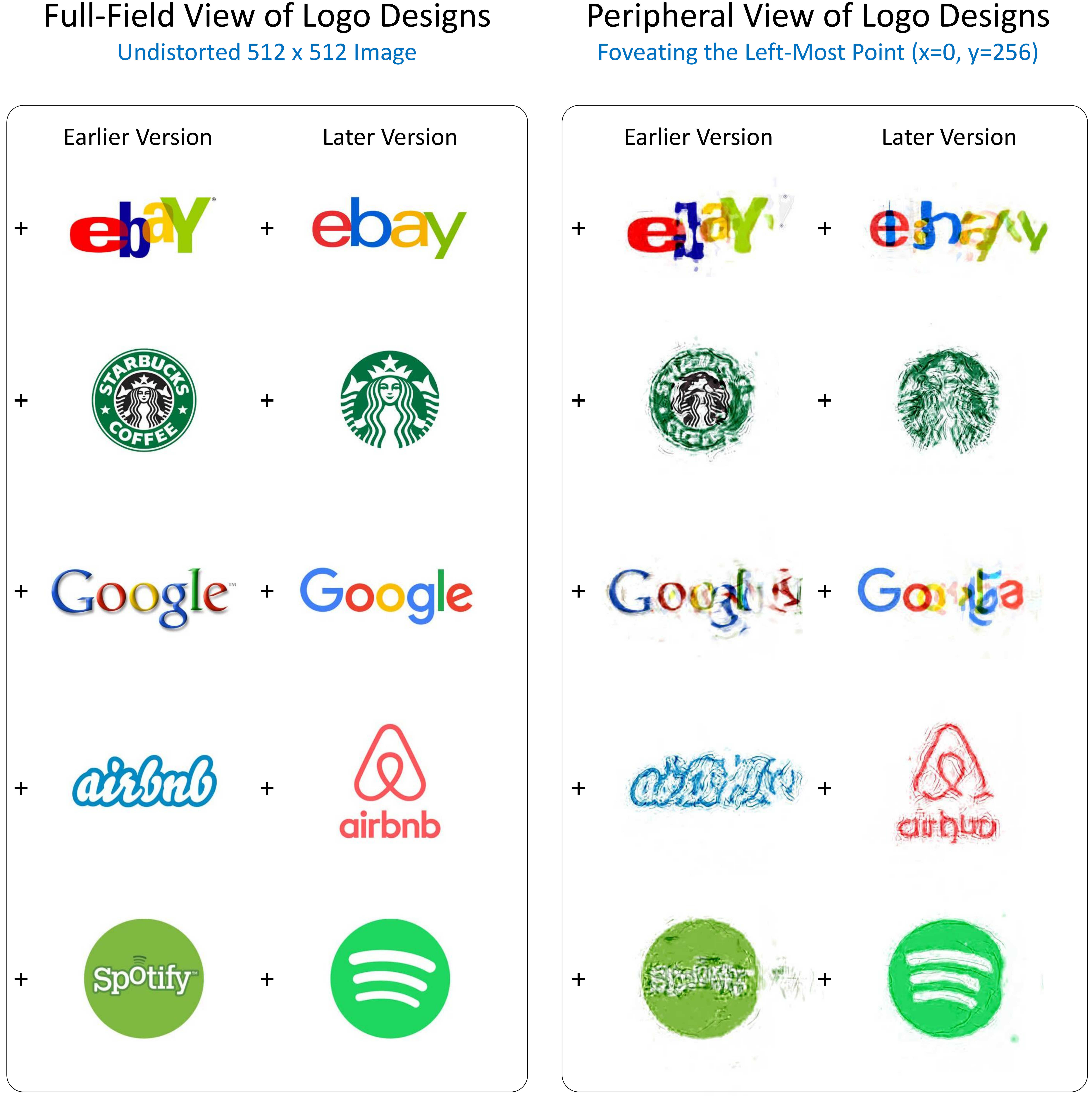}
  \caption{Visualization of how two version of a famous brand's logo appear in the periphery. The left two columns show
    the undistorted full-field views of the logos. The right two columns show peripheral visualizations of those logos
    when the fovea is centered on the black cross to the left of each respective logo. This kind of peripheral
    visualization process can be used as a tool for understanding the initial visual exposure to a design when it first
    appears in the observer's periphery.}
  \label{fig:logos}
\end{figure*}

The ability to recognize a logo at a glance, even in the periphery, has significant potential upside for brand exposure
and recall \cite{van2016logo}. We can use peripheral vision simulation to evaluate how quickly the recognizability of a
particular logo design degrades in the periphery. Understanding how the logo appears in the observer's periphery may
help understand the potential effectiveness of product placement in the physical world and in layout design on the
web or on the printed page.

\figref{logos} shows the application of FGN to two versions of famous logos: (1) an earlier version of the company's
logo and (2) the current version of their logo. The left two columns show logos in their undistorted version. The right
two columns visualize these logos as they may appear in the periphery when the fovea is centered at the black cross to
the left of the logo in question.

Using the metric of peripheral recognition, the new Spotify, Airbnb, and Google logos appear to be an improvement, while
both the new eBay and Starbucks logos appear to have a decreased recognizability in the periphery. It should be noted,
that there may be other metrics under which the results are reversed, such as how memorable the logo is on repeated
viewing. In that sense, peripheral vision simulation may be a useful process in a broader analysis of logo design.

\section{Conclusion}

We show that a generative network with a spatial foveating component can learn in an end-to-end way to efficiently
estimate the output of a human-realistic model of peripheral vision. We achieve a 4 orders-of-magnitude decrease in
running time, from 4.2 hours per image to 0.7 seconds per image. This kind of jump in performance opens the door to a
wide variety of applications from interface design to virtual reality to video foveation for studying human behavior and
experience in real-world interactions.

\paragraph{Code, Data, and Future Work}

The SideEye tool and the FGN source code are made publicly available at \periphurl. In addition, the TTM-generated
images used for training are made available online. Future work will extend the size of the TTM dataset from 1,000 to
100,000 images. This will allow other groups to propose better-performing end-to-end architectures trained on the TTM
model.



\clearpage
\bibliographystyle{sigchi}
\bibliography{peripheral}

\end{document}